\documentclass[letterpaper]{article} 
\usepackage{aaai23}  
\usepackage{times}  
\usepackage{helvet}  
\usepackage{courier}  
\usepackage[hyphens]{url}  
\usepackage{graphicx} 
\usepackage{natbib}  
\usepackage{caption} 
\usepackage{algorithm}
\usepackage{algorithmic}

\usepackage{newfloat}
\usepackage{listings}
\DeclareCaptionStyle{ruled}{labelfont=normalfont,labelsep=colon,strut=off} 
\floatstyle{ruled}
\newfloat{listing}{tb}{lst}{}
\floatname{listing}{Listing}
\usepackage{mathtools}
\usepackage{amsfonts}
\usepackage[dvipsnames]{xcolor}
\usepackage{booktabs, multirow} 

\title{Multi-label Few-shot ICD Coding as Autoregressive Generation with Prompt}
\author{
    Zhichao Yang, \textsuperscript{\rm 1}
    Sunjae Kwon,  \textsuperscript{\rm 1}
    Zonghai Yao,  \textsuperscript{\rm 1}
    Hong Yu,      \textsuperscript{\rm 1,2,3}
}
\affiliations{
    \textsuperscript{\rm 1}College of Information and Computer Sciences, University of Massachusetts Amherst\\
    \textsuperscript{\rm 2}Department of Computer Science, University of Massachusetts Lowell\\
    \textsuperscript{\rm 3}Center for Healthcare Organization and Implementation Research, Veterans Affairs Bedford Healthcare System \\

    zhichaoyang@umass.edu
}

\begin{document}

\maketitle

\begin{abstract}
Automatic International Classification of Diseases (ICD) coding aims to assign multiple ICD codes to a medical note with an average of 3,000+ tokens. 
This task is challenging due to the high-dimensional space of multi-label assignment (155,000+ ICD code candidates) and the long-tail challenge -
Many ICD codes are infrequently assigned yet infrequent ICD codes are important clinically. 
This study addresses the long-tail challenge by transforming this multi-label classification task into an autoregressive generation task. 
Specifically, we first introduce a novel pretraining objective to generate free text diagnoses and procedures using the SOAP structure, the medical logic physicians use for note documentation. 
Second, instead of directly predicting the high dimensional space of ICD codes, our model generates the lower dimension of text descriptions, which then infers ICD codes. 
Third, we designed a novel prompt template for multi-label classification.
We evaluate our Generation with Prompt (GP\textsubscript{soap}) model with the benchmark of all code assignment (MIMIC-III-full) and few shot ICD code assignment evaluation benchmark (MIMIC-III-few).
Experiments on MIMIC-III-few show that our model performs with a marco F1 30.2, which substantially outperforms the previous MIMIC-III-full SOTA model (marco F1 4.3) and the model specifically designed for few/zero shot setting (marco F1 18.7). 
Finally, we design a novel ensemble learner, a cross-attention reranker with prompts, to integrate previous SOTA and our best few-shot coding predictions. 
Experiments on MIMIC-III-full show that our ensemble learner substantially improves both macro and micro F1, from 10.4 to 14.6 and from 58.2 to 59.1, respectively. 


\end{abstract}

\section{Introduction}

In real-world tasks, there are often insufficient training data for rare class labels \citep{Atutxa2019InterpretableDL, wilds2021}. 
Taking automatic international classification of diseases (ICD) coding \citep{Larkey1996Combining} as an example, 
given a discharge note as input, the task is to assign multiple ICD disease label codes associated with each note. 
The ICD coding task in MIMIC-III dataset \citep{Johnson2016MIMICIIIAF} contains 4,075 unique ICD-9-CM codes in 3,372 testing data, among which 1,285 (31.5\%) codes occur less than 11 times in the training split. 
In the clinical domain, rare codes may be as clinically important as common codes for a patient. Therefore, a multi-label classifier is required to perform with high precision and recall for every ICD code, including infrequent ones. 



\begin{figure}[t]
	\centering
	\includegraphics[width=0.45\textwidth]{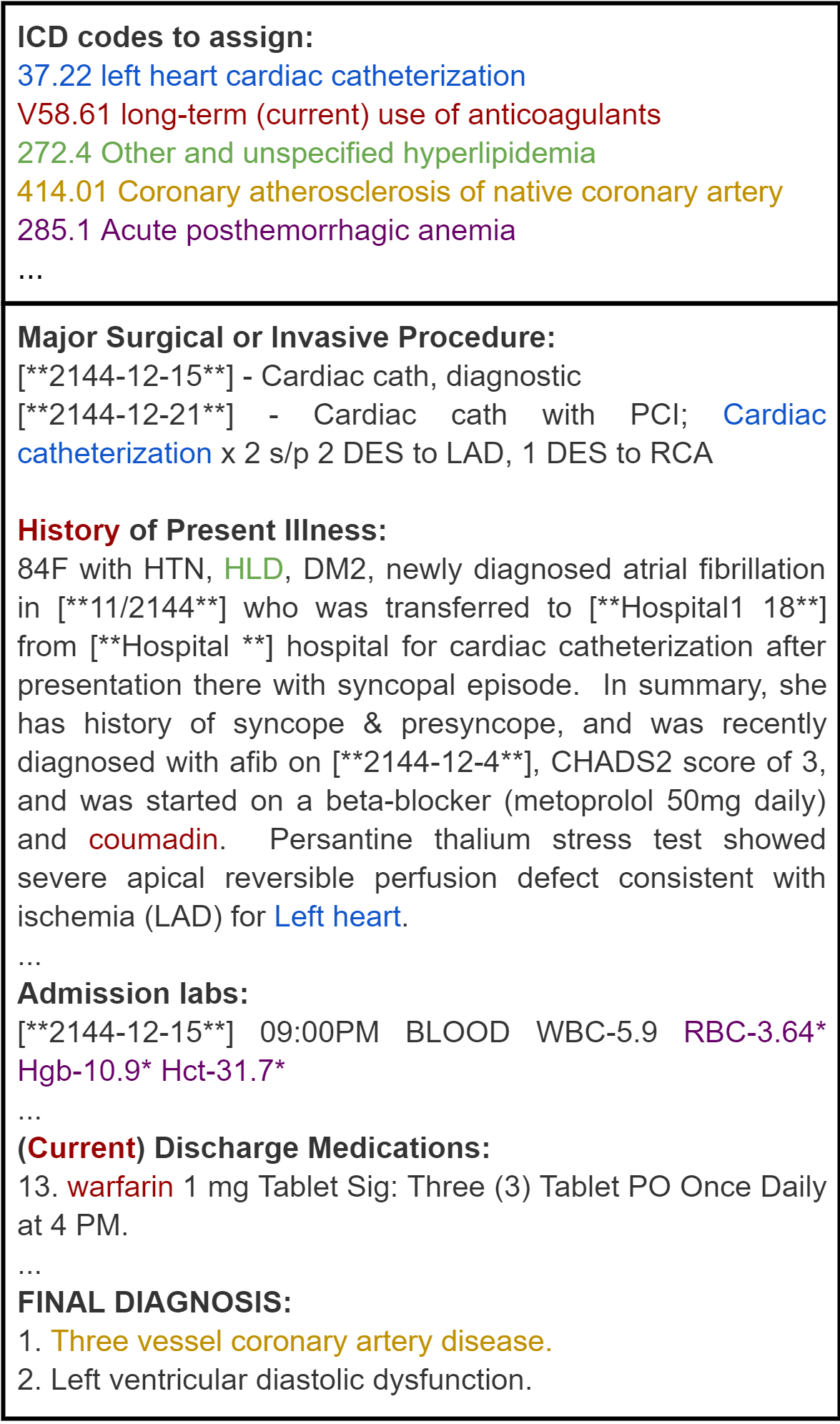}
	\caption{An example of EHR note from the MIMIC-III dataset, which includes ICD code gold label (top) and the discharge note text (bottom). We colored each code and its corresponding candidate mentions/evidence in the note text.}
	\label{fig:example}
\end{figure}

Many recent studies tried self-attention BERT-based models on ICD coding \citep{zhang-etal-2020-bert, Ji2021DoesTM, pascual-etal-2021-towards, Dai2022RevisitingTM}. Most existing BERT-based models follow a common design choice:
each ICD code is associated with a unique entity representation and the coding can be transformed into a multi-label classification across ICD codes. Specifically, previous models use bi-encoder to calculate a low-dimensional dense vector representation for an ICD code and for a note text separately. Later, they take dot product of the two to build a note-aware code representation for each ICD code. 
This design enables a multi-label classification with binary cross-entropy loss on all code representations. However, such a design has several limitations.

First, bi-encoder may miss fine-grained interactions between target ICD codes and candidate mentions in discharge note texts \citep{Mullenbach2018ExplainablePO}. 
A discharge note contains hundreds of different candidate mentions.
Take an ICD code \textcolor{NavyBlue}{\textit{left heart cardiac catheterization}} in Figure \ref{fig:example} as an example, their note encoder aims to capture two candidate mention phrases \textcolor{NavyBlue}{cardiac catheterization} and \textcolor{NavyBlue}{left heart}.
The encoder also needs to repeat this process for every code and condenses the note text into a single low-dimensional dense vector. When pooling thousands of tokens from a note text into a single representation, it is possible that some detailed level of information would be lost. 
This detail-lost representation of note text would then miss fine-grained interactions when cross attend to target code descriptions.

Second, note-aware code representation from cross-attention is unable to generate from missing mention, because language model contains limited amount of medical knowledge \citep{meng-etal-2022-rewire, Yao2022ContextVE}. Take ICD code \textcolor{Plum}{ \textit{Acute posthemorrhagic anemia}} in Figure \ref{fig:example} as an example, there is no mention of \textit{anemia} in this discharge note texts. However, the model is required to infer from lab results such as low level of \textcolor{Plum}{RBC}, \textcolor{Plum}{Hgb}, \textcolor{Plum}{Hct}, which leads to \textit{anemia}. In order to fill in this knowledge gap, previous researches use additional information such as hierarchical code ontology \citep{tsai-etal-2019-leveraging, cai-etal-2022-generation}, code co-occurrence \citep{Cao2020HyperCoreHA}, code frequency \citep{zhou-etal-2021-automatic}, signs and symptoms section in Wikipedia \citep{ijcai2020-471, wang-etal-2022-novel} as an additional input source, and use the same binary cross-entropy objective on codes. However, few research design a specific training objective to distill knowledge into their model.

\begin{figure*}[t]
	\centering
	\includegraphics[width=0.98\textwidth]{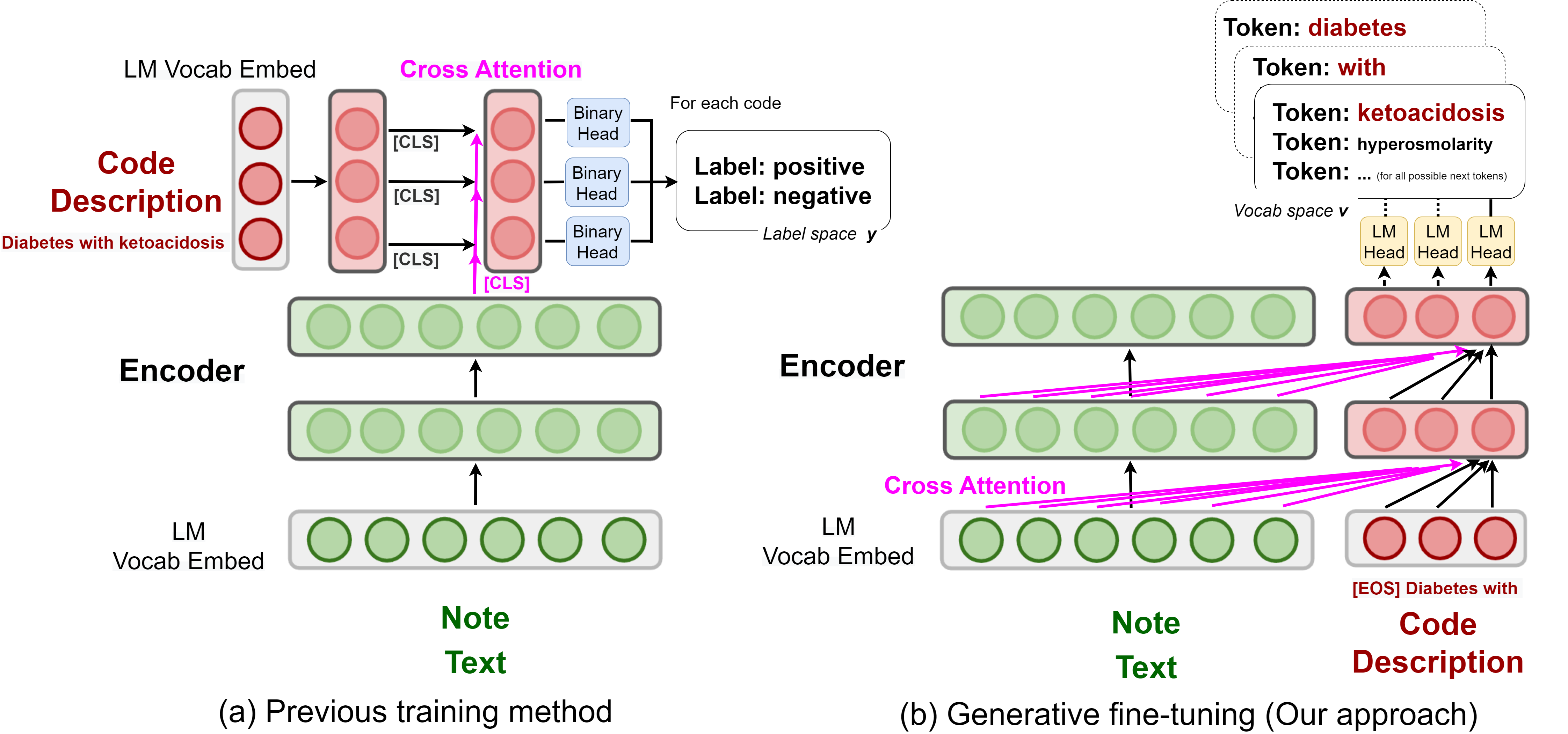}
	\caption{An illustration of (a) standard training method and (b) our proposed generative fine-tuning. The number of candidate code is 3 for illustration in (a). Solid lines represent attention in step 3 and dash lines represent previous steps in (b).}
	\label{fig:model}
\end{figure*}

Besides the accuracy challenge from the long tail distribution, such a high number of ICD codes also increases their model's memory demand.
As stated previously, for the ICD-9 testing data from MIMIC-III, the model needs to assign from 4,075 unique diagnosis and procedure codes. In reality, ICD-9 has a total of 14,000 and 3,900 unique diagnosis and procedure codes, respectively. The number of unique ICD codes is also increasing because more and more diseases are being refined or added. The up-to-date ICD-10, which replaces ICD-9, contains 68,000 diagnosis codes and 87,000 procedure codes \citep{Eisfeld2014InternationalSC}. Most previous methods require building a note-aware code representation for each code as shown in Figure \ref{fig:model}. The number of code representations stored in memory increases linearly as the number of candidate codes to assign increases. The significant increase in memory makes it hard to deploy in the real-world auto ICD classification setting \citep{Ziletti2022MedicalCW, YAN2022}.

In this paper, we present a simple and yet effective autoregressive Generation with Prompt (GP) framework for ICD coding. 
We have made major contributions to ICD coding. Specifically, we designed a novel mask objective for the clinical language model. Our pretraining objective utilizes the unique clinical knowledge inherited by the structure of EHR notes. Specifically, physicians write notes following the subjective, objective, assessment, and plan (SOAP) structure, where the assessment and plan sections can be inferred from the subjective and objective sections \cite{umls2004, yang-yu-2020-generating}. 
Our GP\textsubscript{soap} is a longformer encoder-decoder (LED) pretrained with assessment \& plan generation loss, which infers diagnoses and procedures from symptoms and lab results by generating the assessment and plan sections from the subjective and objective sections. This pretraining can mitigate the missing mention challenge.

In addition, unlike all previous approaches which directly predict ICD codes, we fine-tune GP\textsubscript{soap} to first generate code descriptions with a prefix prompt and then assign the descriptions to the corresponding ICD codes. Code description generation only requires a fixed vocab of 4,501 candidate word tokens, a substantial reduction in dimensional space.
Unlike all previous models that build cross-attention on a single pooled representation of input note texts, we perform an autoregressive cross-attention on every single token of EHR notes to avoid the problem of missing fine-grained interactions. 

Experiments on the public ICD coding benchmark (MIMIC-III-full) show that our model GP\textsubscript{soap} outperforms the state-of-the-art (SOTA) MSMN on MIMIC-III-full using the macro evaluation metrics.
In order to verify its few-shot capability, we also show that our model substantially outperforms both MSMN \citep{Yuan2022CodeSD} and AGMHT \citep{Song2020GeneralizedZT}. 
Last but not least, we show that an ensemble learner which simply reranks the predictions of our model and MSMN achieves the new SOTA on MIMIC-III-full.
Our codes are attached in supplementary material and will be publicly available upon publication.

\section{Methods}

\subsection{Task Definition}
ICD coding is typically formulated as a multi-label classification task \citep{McCallum99multi-labeltext}.
Specifically, considering thousands of words from an input \textcolor{ForestGreen}{discharge summary note text} \textcolor{ForestGreen}{$t$} (green as shown in Figure \ref{fig:model}), the task is to assign a binary label $y_i \in \{0,1\}$ for the $i$th ICD code in the label space $Y$, where 1 means that note is positive for an ICD diagnosis or procedure. 
Each candidate code has a short code description phrase $c_i$ in free text. For instance, code 250.1 has description \textit{diabetes with ketoacidosis}. \textcolor{BrickRed}{Code descriptions} \textcolor{BrickRed}{$c$} is the set of $N_c$ number of $c_i$ where $N_c=4,075$ is the number of total ICD codes unless otherwise specified.

\subsection{Pretraining to Learn Clinical Inference} \label{sec:pretrain}
Besides token-level masked language modeling (MLM) objective, which has shown to be limited in biomedical tasks \citep{Moradi2021GPT3MA, Gutierrez2022ThinkingAG}, our GP\textsubscript{soap} model is also pretrained with a paragraph-level autoregressive encoder-decoder objective. 
Specifically, given subject and object (SO) sections including symptoms and lab results, the task is to generate free-text assessment\&plan (AP) including diagnoses and procedures (refer to figure 1 of \citet{yang-yu-2020-generating} for an example\footnote{https://aclanthology.org/2020.findings-emnlp.336.pdf}). 
The SOAP is identified using the document section classification tool \citep{Sunjae2022autoSOAP}.
Similar to PEGASUS \citep{Zhang2020PEGASUSPW}, we masked selected sentences in assessment\&plan section with [MASK1] for target generation text (GSG). 
The selected sentences include diagnoses and procedures entities found using biomedical named entity recognition and linking (NER+L) tool MedCAT \citep{Kavuluru2015AnEE}.
The other sentences in subject and object section remain almost the same as input, but with some phrases randomly masked by [MASK2] for MLM. This pretraining objective is similar compared to ICD coding task, which aims to assign ICD diagnoses and procedures codes, but differs from ICD coding in that it generates sentences other than the specific code. ICD coding also has the whole medical note as input, but our GSG only includes part of clinical notes without diagnoses and procedures. Hence, our GP\textsubscript{soap} could learn such inferences from symptoms and lab results to diagnoses and procedures.

\subsection{Generative Fine-tuning} \label{sec:our_method1}
Our generative fine-tuning is different from standard fine-tuning. 
We use autoregressive encoder-decoder model to assign ICD codes to a note by first generating unique code descriptions. An example to assign the ICD code $250.1$ with the corresponding text description \textit{diabetes with ketoacidosis} is illustrated in Figure \ref{fig:model} (b). This step is repeated $N$ times for each code description to generate, and PLM implicitly decides $N$ by generating the end of text token. 
In this way, we transfer a downstream multi-label classification task into a sequence to sequence task like pretraining.
Specifically, we design a free text prompt template as input:

$x_i =$ \textcolor{ForestGreen}{t}. diagnoses and procedures : \textcolor{BrickRed}{$c_1$} ; \textcolor{BrickRed}{$c_2$} ;  ... ; \textcolor{BrickRed}{$c_{i}$} . 

Given $i$th prompt, the next $i+1$th code label probability would be calculated as:

\begin{equation}
\begin{aligned}
P(\hat{c}_{i+1} | x_{i}) &=  \prod_{j=1}^{L_c} P(u_{j} | u_{<j}, x_{i})
\end{aligned}
\label{eq:mPLM}
\end{equation}

\noindent where u is the sequence of $L_c$ tokens in $i+1$th code description $\hat{c}_{i+1}$ to generate. 
Finally, we grouped the generated tokens as a code description phrase $\hat{c}_{i+1}$, which is then mapped to the corresponding ICD code. During prediction, the ICD code is assigned with the binary label of $y=1$, and use the same code to evaluate and compare to standard fine-tuning.
Notice that this decoding step would \textcolor{CarnationPink}{cross-attend} each token of \textcolor{BrickRed}{code descriptions} and \textcolor{ForestGreen}{note text} in every layer of PLM decoder. Generative fine-tuning reuses all parameters during pretraining, and does not introduce new randomly initialized parameters, making the whole model easy to fine-tune in a few-shot setting.

\subsection{Inference with Clinical Vocab Constraint} \label{sec:our_method2}
\begin{figure}[t]
	\centering
	\includegraphics[width=0.46\textwidth]{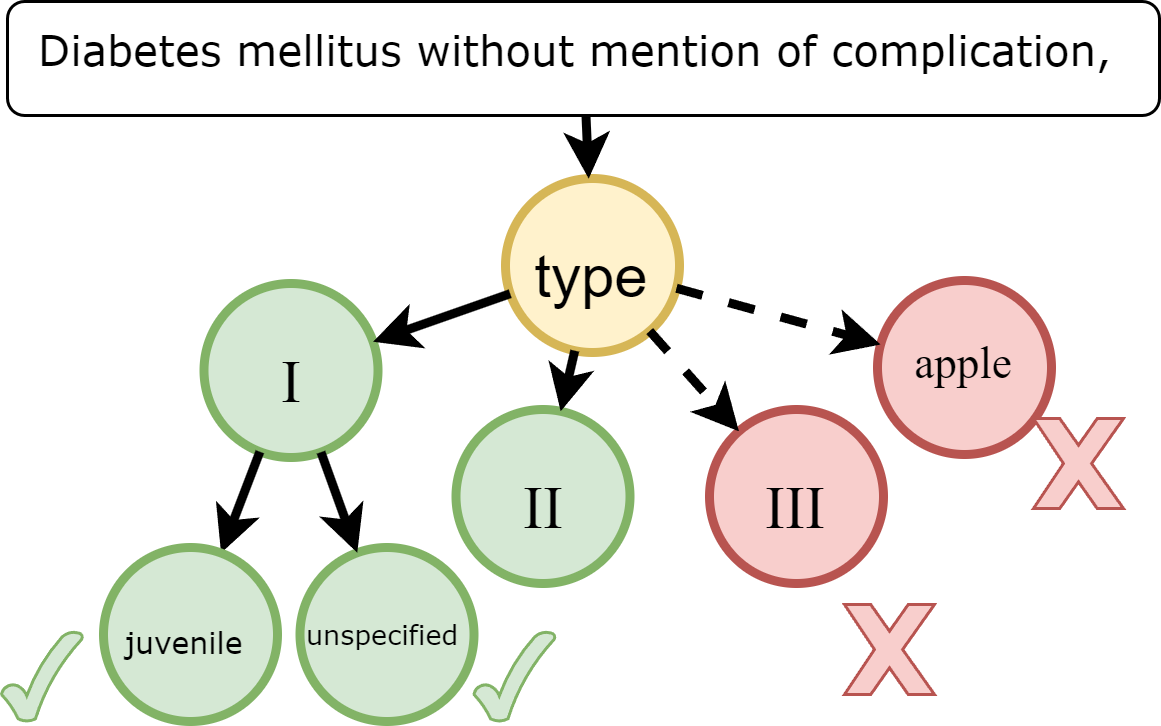}
	\caption{An illustration of trie constraint beam search which forces to decode only valid code descriptions. Full valid description is \textit{Diabetes mellitus without mention of complication, type 1 juvenile type, not stated as uncontrolled}.}
	\label{fig:trie}
\end{figure}

Ideally, the generated tokens would match to a valid code. However in reality, PLM would hallucinate and generate random tokens that do not match to any code descriptions, especially given that code description is usually a long phrase of average 9.81 tokens. 
Hence, we exploit beam search \citep{NIPS2014_a14ac55a} with vocab constraint, which forces to only generate valid code descriptions. Similar to previous research on entity retrieval \citep{decao2021autoregressive}, we applied a trie structure to efficiently limit the vocab. Since beam search only examines one step forward during decoding, we only limit the generation of one future token given the preceding tokens. Hence, we construct the trie by the following rule: for each node in the trie, its children are allowed next tokens to formulate into a valid code description. While querying the allowed next tokens for a sub phrase, we simply traverse from root node using tokens in sub phrase to find the children. An illustrative example is shown in Figure \ref{fig:trie}.

By querying this trie, we are able to generate a valid description for a single code. In order to generate multiple codes, we design 2 different ending tokens for each code description: [EOS] (end of text token) or ; (semicolon), where [EOS] means this is the last code to assign to this note text, and semicolon means continue generating other codes. We let decoder decide how many codes to assign.

\subsection{Code Reranking}
Instead of generative prompt intended for few-shot learning, we propose a novel cloze style prompt to rerank predicted code candidates generated from different models. 
The concept of 2nd stage reranker originates from information retrieval \citep{dang2013two, Nogueira2019PassageRW}.
This reranker serves as a fine-grained coding model. Specifically, we reformulate multi-label classification tasks with free text prompt template as input: 

$x_p =$\textcolor{ForestGreen}{t} . \textcolor{BrickRed}{$c_1$} : [MASK] , \textcolor{BrickRed}{$c_2$} : [MASK] , ... ,  \textcolor{BrickRed}{$c_{N_b}$} : [MASK] .   

\noindent where $c_i$ is the descriptions of code candidates generated from previous step models. We use KEPTLongformer \citep{Yang2022KnowledgeIP} to decide if code is positive (or negative) for a note by filling [MASK] with vocab token yes (or no). This step is very similar to generative fine-tuning but differs in the design of the prompt. Instead of using the prompt to autoregressively generate the next code description, our reranker formulates the prompt with the code description given as the candidate, and verifies if the code should be assigned to the discharge note texts or not.

\section{Experiments}

\subsection{Dataset}
To reduce potential bias with small cohort\citep{Weber2017BiasesIB}, we pretrain on a large cohort containing 3.6 million clinical notes following SOAP structure. The pretraining cohort is from US Veterans Health Administration Corporate Data Warehouse. We collect every patient progress notes from calendar year 2016 to 2020. 
The average number of tokens for SO section and AP section is 1456 and 175 respectively.

The fine-tuning dataset \citep{Johnson2016MIMICIIIAF} contains clinical data from real patients. It contains data instances of de-identified discharge summary note texts with expert-labeled ICD-9 codes. We applied the following text pre-processing steps before tokenization: (1) removing all de-identification tokens; (2) replacing characters other than punctuation marks and alphanumerical into white space (e.g. /n); (3) stripping extra white spaces. The average number of documents per label is 12.6, and the average labels per document is 15.8.

For all codes prediction tasks (MIMIC-III-full), we used the same splits as the previous work \citep{Mullenbach2018ExplainablePO, Yuan2022CodeSD}. Previous work truncated discharge summaries at 4,000 words. Since longformer used tokens instead of words, we truncated discharge summaries at 8,192 tokens unless otherwise specified.

To benchmark ICD coding task on few shot learning, we also created a few-shot codes prediction dataset (MIMIC-III-few), which is a subset of the MIMIC-III-full dataset. Following previous works on few-shot classification \citep{Lu2020MultilabelFL,Song2020GeneralizedZT}, among 4,075 different types of ICD-9 codes in test set, we collect those which occur within 5 times (but occur at least 1 time) in the training set. We found a total of 685 (16.8\%) unique ICD-9 codes under the few-shot setting.

\subsection{Baselines}

\noindent \textbf{MSMN} \citep{Yuan2022CodeSD}
use synonyms with adapted multi-head attention, which achieved SOTA performance on the MIMIC-III-full dataset.

\noindent \textbf{AGMHT} \citep{Song2020GeneralizedZT}
use adversarial generative model to generate feature for zero/few shot labels, which achieved SOTA performance on zero-shot MIMIC-III task.

\noindent \textbf{GP\textsubscript{wiki}}
use our GP framework (without GSG pretraining objective in section \ref{sec:pretrain}). During fine-tuning, it is initialized from LED pretrained on Wikipedia and Book corpus in \citet{Beltagy2020LongformerTL}.

\noindent \textbf{GP\textsubscript{pubmed}}
use our GP framework but initialized from LED pretrained on the Pubmed dataset.

\begin{table*}
\centering
\renewcommand{\arraystretch}{1.2}
\begin{tabular}{lrrrrrrrrrrr}\toprule
\multirow{2}{*}{Model} &\multicolumn{2}{c}{F1} &\multicolumn{2}{c}{Prec} &\multicolumn{2}{c}{Recall} & & & & \\\cmidrule{2-7}
&Mac &Mic &Mac &Mic &Mac &Mic &P@15 &R@15 &P@50 &R@50 \\\midrule
Standard  &7.7 &47.2 &7.6 &42.4 &7.7 &53.4 &47.2 &43.2 &24.9 &76.1 \\
MSMN &10.4 &\textbf{58.1} &10.4 &\textbf{61.2} &10.5 &\textbf{55.4} &\textbf{59.6} &\textbf{55.8} &\textbf{27.0} &\textbf{78.4} \\
AGMHT &1.3 &0.6 &1.2 &36.6 &1.5 &0.3 &- &- &- &- \\
GP\textsubscript{pubmed} &8.3 &44.2 &10.1 &54.7 &7.1 &37.0 &- &- &- &- \\
GP\textsubscript{soap} &\textbf{13.4} &49.8 &\textbf{13.2} &53.7 &\textbf{13.7} &46.4 &- &- &- &- \\
\hline
Reranker Oracle &28.8 &67.9 &28.2 &63.7 &29.4 &72.7 &63.6 &59.0 &35.9 &99.8 \\
Reranker (MSMN+AGMHT) &11.9 &\textbf{59.7} &12.5 &\textbf{65.3} &12.2 &57.5 &\textbf{61.4} &\textbf{57.4} &27.0 &78.6 \\
Concater (MSMN+AGMHT) &11.6 &58.2 &11.4 &61.0 &11.9 &55.6 &- &- &- &- \\
Reranker (MSMN+GP\textsubscript{pubmed}) &11.7 &57.2 &12.2 &57.9 &11.2 &56.5 &58.7 &55.5 &27.0 &78.5 \\
Concater (MSMN+GP\textsubscript{pubmed}) &11.4 &54.8 &11.0 &52.6 &12.6 &59.4 &- &- &- &- \\
Reranker (MSMN+GP\textsubscript{soap}) &\textbf{14.6} &59.1 &\textbf{14.3} &58.3 &14.9 &59.9 &60.5 &56.8 &\textbf{27.8} &\textbf{80.2} \\
Concater (MSMN+GP\textsubscript{soap}) &14.0 &55.0 &12.0 &50.2 &\textbf{17.6} &\textbf{60.8} &- &- &- &- \\
\bottomrule
\end{tabular}
\caption{Results on the MIMIC-III-full set, compared between GP and baselines (top), reranking from combination of different baselines (down). Standard is standard fine-tuning with longformer as encoder. AGMHT is only built for zero/few-shot.}\label{tab:result_full}
\end{table*}

\subsection{Metrics}
We report both macro and micro scores of precision, recall, f1, and accuracy. Given the predicted text code pair, macro-averaged values are calculated by first computing metric for each ICD code and then taking the average among all classes.
Micro-averaged values are calculated by treating each pair as a separate prediction and average among all predictions. 
Take precision as an example, the metrics are distinguished as follows:

\begin{equation}
\begin{aligned}
Macro P = 1/ N_c \sum_{i=1}^{N_c} \frac{TP_i}{TP_i+FP_i} \\
\end{aligned}
\label{eq:macrop}
\end{equation}

\begin{equation}
\begin{aligned}
Micro P = \frac{\sum_{i=1}^{N_c} TP_i}{\sum_{i=1}^{N_c} TP_i+FP_i} \\
\end{aligned}
\label{eq:microp}
\end{equation}

\noindent where $TP_i$ is True Positives for code label $i$, and $FP_i$ is False Positives for code label $i$. This shows Macro metrics place much more emphasis on rare code prediction. 

We also report precision or recall at K. This is inspired by the need of clinical decision support tool when applied to real world, in which a doctor is given a certain number of anticipated codes to examine.
We select $K=15$ which is roughly the average number of codes in MIMIC-III per note texts, and $K=50$ as this is the maximum candidate our largest GPU memory support as a reranker.

\subsection{Implementation Details}
We order ICD codes to generate by its SEQ\_NUM (priority ranking for diagnoses and perform order for procedures) labeled by medical experts from DIAGNOSES\_ICD PROCEDURES\_ICD table in MIMIC-III. For punctuation semicolons in ICD code descriptions, we replace them to punctuation commas. 
Pretraining on SOAP data took about 140 hours with 4 NVIDIA RTX 6000 GPU with 24 GB memory. Fine-tuning took about 40 hours with 4 NVIDIA RTX 6000 GPU with 24 GB memory. Our reranker training took about 12 hours with 2 NVIDIA A100 GPU with 40 GB memory. During pretraining, we used warmup ratio of $0.1$, learning rate ${5e^{-5}}$, dropout rate $0.1$, L2 weight decay $1e^{-3}$ and batch size of 64 with fp16. During fine-tuning, we grid searched learning rate [$1e^{-5},\mathbf{2e^{-5}},3e^{-5}$], dropout rate [$\mathbf{0.1},0.3,0.5$], with batch size of 4. Best hyper-parameters set is bolded. For each evaluation on generation models, we run generation once with a beam search size of 2. We set the number of candidate code to rerank $N_b = 50$ as this is the maximum candidate our largest GPU memory support. Our evaluation code is publicly available\footnote{\url{https://github.com/whaleloops/KEPT}}.

\subsection{Results} \label{sec:results}
Results show that our generation with prompt framework pretrained on clinical SOAP data (GP\textsubscript{soap}) migrates the long-tail issue compared to the previous SOTA model MSMN.
For the all disease code assignment (MIMIC-III-full) task, our GP\textsubscript{soap} achieves macro F1 of 13.4 (+3.0), macro precision of 13.2 (+2.8), and macro recall of 13.7 (+3.2). The number in parentheses shows the improvements compared to MSMN. For the rare disease code assignment (MIMIC-III-few) task, our GP\textsubscript{soap} achieves macro F1 of 30.2 (+25.9), micro F1 of 35.3 (+26.8), macro recall of 32.9 (+28.7), micro recall of 32.6 (+28.1), macro accuracy of 25.1 (+21.3), micro accuracy of 21.4 (+17.0). This further confirms the strong advantage of our GP\textsubscript{soap} for long-tail codes. 
Finally, by combining the prediction of GP\textsubscript{soap} and MSMN, our reranker with code descriptions as prompt shows SOTA performance for both macro and micro evaluations.

\begin{table}
\centering
\begin{tabular}{lrrrrrrr}\toprule
\multirow{2}{*}{Model} &\multicolumn{2}{c}{F1} &\multicolumn{2}{c}{Prec} &\multicolumn{2}{c}{Recall} \\\cmidrule{2-7}
&Mac &Mic &Mac &Mic &Mac &Mic \\\midrule
MSMN &4.3 &8.5 &4.5 &\textbf{70.9} &4.2 &4.5 \\
AGMHT &18.7 &29.2 &17.6 &49.4 &19.9 &20.7 \\
GP\textsubscript{wiki} &2.9 &4.9 &3.0 &61.1 &2.7 &2.6 \\
GP\textsubscript{pubmed} &7.3 &12.3 &7.9 &47.3 &6.8 &7.1 \\
GP\textsubscript{soap} &\textbf{30.2} &\textbf{35.3} &\textbf{27.9} &38.5 &\textbf{32.9} &\textbf{32.6} \\
\bottomrule
\end{tabular}
\caption{Results on the MIMIC-III-few set, compared between baselines and different pretraining of GP.}\label{tab:result_few} 
\end{table}


\section{Discussions}
Our final model is a combination of 1) autoregressive generation; 2) SOAP pretraining objective; 3) reranker with prompt. We will discuss details in the following paragraphs:

\textbf{Are autoregressive coding models few-shot learners?}

\noindent Our autoregressive model GP\textsubscript{soap} outperforms non-autoregressive auto coding model MSMN in few-shot coding task as shown in Section \ref{sec:results}. Moreover, our GP\textsubscript{soap} also significantly outperforms AGMHT which is specifically designed for few/zero-shot ICD coding. Compared to AGMHT, our GP\textsubscript{soap} improves macro and micro F1 by +11.5 and +6.1, macro and micro precision by +10.3, -10.9, macro and micro recall by +13.0, +11.9, macro and micro accuracy by +9.2, +4.3. Our findings support that clinical PLMs are few shot learners \citep{Taylor2022ClinicalPL, lewis-etal-2020-pretrained, Yang2022GatorTronAL}. 
From a network architecture perspective, our autoregressive ICD generation model uses cross-attention to capture note aware code representations, whereas previous traditional ICD coding studies use dot product label attention \citep{zhang-etal-2020-bert, Ji2021DoesTM, pascual-etal-2021-towards, Dai2022RevisitingTM}.
A dot product label attention is required to aggregate many tokens from note into a single note representation before a final label cross-attention layer to build note aware code representations. Such a selection of which tokens best represent note is training intensive and does not generalize well in out-of-domain or zero/few-shot setting \citep{NEURIPS2020_3f5ee243,Thakur2021BEIRAH}.
Traditional methods avoid this by using a label attention when creating note aware code representations \citep{Mullenbach2018ExplainablePO, Bai-etal-2019-improving, ji-etal-2020-dilated, wangCoding2020, zhou-etal-2021-automatic, liu-etal-2021-effective, Kim2021ReadAA, luo-etal-2021-fusion, Sun2021MultitaskBA, wang-etal-2022-novel, Yuan2022CodeSD, Ren-etal-HiCu}. However, the majority of them used unpretrained LSTM or CNN. This increased the number of unpretrained parameters and has shown to limit its ability in few-shot setting \citep{NEURIPS2020_1457c0d6,schick-schutze-2021-just,gao-etal-2021-making}.
In contrast, our model combines the merit of both PLM and traditional ICD coding model. Our model is pretrained without introducing new unpretrained parameters during fine-tuning and also cross-attend code with every token in the note like traditional label attention.
Such cross-attention has shown to be effective (compared to a typical bi-encoder network) in combining query and document in information retrieval passage ranking \citep{zhou-devlin-2021-multi}, but has not yet been deployed for multi-label classification tasks. 

\textbf{Comparison among different pretraining objectives in adding clinical knowledge into PLM}.

\begin{table}
\centering
\begin{tabular}{lrrrrr}\toprule
\multicolumn{2}{c}{MIMIC-III-few} &F1 &Prec &Recall \\\midrule
\multirow{2}{*}{GP\textsubscript{pubmed}} &Mentioned &17.3 &17.9 &17.9 \\
&UnMentioned &7.0 &8.6 &6.7 \\
\multirow{2}{*}{GP\textsubscript{soap}} &Mentioned &21.3 &17.2 &35.5 \\
&UnMentioned &17.1 &15.0 &25.2 \\
\bottomrule
\end{tabular}
\caption{The impact of whether diagnoses are mentioned (or not) in discharge summaries. GP\textsubscript{soap} is pretrained with GSG objective while GP\textsubscript{pubmed} is not. }\label{tab:result_mentioned} 
\end{table}

\noindent In-domain pretraining is important for clinical PLM to perform well in few-shot setting. 
Hence, we designed a new training objective on the clinical notes. Instead of the traditional masked language model which captures linguistic features, we designed a novel assessment\&plan generation objective that focuses on predicting clinical outcomes in tokens. Our GP\textsubscript{soap} outperforms GP\textsubscript{pubmed} in recall and F1 scores in both MIMIC-III-few and MIMIC-III-full.
This is mainly due to the learned inference ability during pretraining where the diagnosis may be missing.
We want to know if GP\textsubscript{soap} is able to assign ICD codes when a missing mention occurs (e.g. \textit{anemia} in Figure \ref{fig:example}.).
We annotate the discharge summaries with ICD-9 diagnosis codes with MedCat \citep{Kavuluru2015AnEE}.
We then evaluate on codes that are explicitly mentioned in the text and those that are not.
Table \ref{tab:result_mentioned} shows that the accuracy on unmentioned diagnoses does not drop significantly from mentioned diagnoses using our GP\textsubscript{soap}, especially when it is compared to GP\textsubscript{pubmed} without assessment\&plan generation objective. The accuracy differences between the two models indicate that assessment\&plan generation objective helps learn more clinical inference ability.


From the view of pretraining objective, our work is most similar to gap sentences generation objective such as PEGASUS \citep{Zhang2020PEGASUSPW}, which has shown to be effective in medical paper summarization \citep{guo-etal-2022-longt5} and healthcare question summarization \citep{Yadav2022CHQSummAD}. \citet{wan-bansal-2022-factpegasus} augment the sentence selection strategy when pretraining and shows improvement in the factuality of the generated summary. Similarly, instead of random text format on MIMIC-III, we select paragraphs following the SOAP format from real-world data, because assessment and plan paragraphs in SOAP contain key clinical information of a patient encounter \citep{stupp-etal-2022-ap}.

From the view of adding medical knowledge into language models,
besides applying masked language modeling which captures linguistic features in the medical domain \citep{Li2019FineTuningBE, Rongali2020Continual}, previous work has integrated knowledge from unlabelled clinical documents through clinical outcome pretraining \citep{van-aken-etal-2021-clinical}, patient journey prediction \citep{peng-etal-2021-self}, amplified vocabulary \citep{Wada2020APT}, and medical entity retrieving \citep{meng-etal-2022-rewire} in non-autoregressive BERT. Here we provide an alternative way of incorporating clinical inference knowledge into an autoregressive model.

\textbf{Combining few-shot and non-few-shot using cross-attention reranker with prompt}.

\noindent Since AGMHT and our pretrained generation model GP\textsubscript{soap} improve the accuracy in rare codes but not for common codes, we further collect the ICD prediction of these models and rerank them with cross-attention prompt based fine-tuning. Compared to MSMN only model, our MSMN+GP\textsubscript{soap} reranker achieves the best macro F1 of 14.6 (+4.2), marco precision of 14.3 (+3.9), macro recall of 14.9 (+4.5), and micro recall of 59.9 (+4.6). MSMN+AGMHT reranker achieves the best micro F1 of 59.7 (+1.5) and micro precision of 62.0 (+0.7). Both models achieve the new SOTA macro and micro F1 on MIMIC-III coding benchmark. We also compare our reranker to a simple concater which directly concatenates predictions of two models. As shown at the bottom of Table \ref{tab:result_full}, all rerankers outperform the corresponding concaters in terms of F1. 
Finally, our final result with reranker shows better F1 micro and macro score compared to RAC \citep{Kim2021ReadAA}, which already reached past the human-level coding baseline.

\section{Related Work}

\subsection{Prompt-based fine-tuning}
Prompt-based fine-tuning has been shown to be effective in few-shot tasks \citep{le-scao-rush-2021-many, gao-etal-2021-making}, even when PLM is relatively small \citep{schick-schutze-2021-just} because they introduce no new parameter during few shot fine-tuning. 
However, most previous works focus on single-label multi-class classification task such as sentiment classification \citep{gao-etal-2021-making}, clinical ICD-9 triage \citep{Taylor2022ClinicalPL}, disease phenotyping \citep{Sivarajkumar2022HealthPromptAZ}.
To the best of our knowledge, our GP is the first work that applies prompting to multi-label classification task.

\subsection{Autoregressive entity linking}
ICD coding is similar to entity linking, as it is the task of assigning a unique textual entity identifier (i.e. entity name) to an entity mention given context. In contrast, ICD coding assigns multiple codes to whole discharge note document.
\citet{decao2021autoregressive} first addresses entity linking by using autoregressive sequence-to-sequence transformer to generate identifier, and experiments show high micro F1 score on both in-domain and out-of-domain entity linking benchmarks. 
\citet{yuan-etal-2022-generative} further enhance autoregressive entity linking using biomedical knowledge base guided pretraining, and achieve SOTA results on biomedical entity linking such as BC5CDR.
Instead of pretraining using a pair of entities within a sentence, we used a pair of subjective-objective paragraphs and assessment-plan paragraphs within SOAP notes each containing multiple entities.

\section{Limitations}

Our best MSMN+GP\textsubscript{soap} has limited accuracy in candidate code assignment, and still has room for improvement compared to oracle.
An oracle candidate code assigning model is able to include all ground truth labels within its top-50 predictions (R@50 is 100).
If such a near oracle model exists, then our reranker is able to achieve macro F1 of 28.8 and micro F1 of 67.9, where our current SOTA only has R@50 of 80.2 and thus achieves lower macro F1 of 14.6 and micro F1 of 59.7. One potential exploration is to change MSMN into more recent ICD coding works like PLM-ICD \citep{huang-etal-2022-plm} which shows higher micro.

Our GP\textsubscript{soap} is pretrained on 3.6 million proprietary clinical notes under SOAP format. We cannot pretrain on publicly available dataset mimic, because mimic discharge summary contains less than 1,000 notes under SOAP format, and thus is not suitable for pretraining. GP is an autoregressive generation method, and hence, slow as it generates one token at a time. In ICD coding task, our generative fine-tuning is about 24 times slower than the non-autoregressive standard fine-tuning.

\section{Conclusions}
In this paper, we investigate a pretrained clinical language model on ICD coding task for both full and rare codes. Adapted from recent advances in autoregressive entity linking, our SOAP pretrained autoregressive model achieves a competitive performance over SOTA systems in rare ICD coding task.
This autoregressive model also preserves the order of the code. In contrast, previous auto ICD coding models neglect this potential error source but are actually common in the real world \citep{Omalley2005MeasuringDI}.
We further propose a cross-attention reranker to combine best prediction from common ICD coding task and rare ICD coding task. Experiments show that our reranker model significantly outperforms the previous SOTA model on full ICD coding task. Our reranker can be easily plugged-in to rerank predictions generated from any model.

\section*{Acknowledgements}
We are grateful to the UMass BioNLP group, especially Avijit Mitra, Bhanu Rawat, David Levy, and Adarsha Bajracharya for many related discussions. We would also like to thank the anonymous reviewers for their insightful feedback. Research reported in this study was supported by the National Science Foundation under award 2124126. The work was also in part supported by the National Institutions of Health R01DA045816 and R01MH125027. The content is solely the responsibility of the authors and does not necessarily represent the official views of the National Science Foundation and National Institutes of Health.

\bibliography{aaai23}


\end{document}